%% file: josis-jury-2013.tex
\documentclass{josis}

\usepackage{amssymb,graphicx,xspace}
\usepackage{multirow}
\usepackage{amsmath}
\usepackage[acronym,xindy]{glossaries}

\usepackage{floatrow}
\usepackage{array}
\usepackage{url}
\usepackage[strings]{underscore}
\usepackage{arydshln}
\newcolumntype{P}[1]{>{\raggedright\arraybackslash}p{#1}}

\newcommand{\urlfoot}[1]{\footnotemark\footnotetext{\url{#1} (accessed on 24/1/2012)}} 

\usepackage[numbers]{natbib} 
\graphicspath{{figures/}}

\input{macros.tex}

\input{glossary.tex}

\runningauthor{\begin{minipage}{.9\textwidth}\centering Ballatore, Bertolotto, Wilson\end{minipage}}
\runningtitle{The Semantic Similarity Ensemble}

\begin{document}

\title{The Semantic Similarity Ensemble\footnote{This article extends work presented at the 6th International Workshop on Semantics and Conceptual Issues in Geographical Information Systems (SeCoGIS 2012) \cite{Ballatore:2012:jury}}}

\author{Andrea Ballatore}\affil{School of Computer Science and Informatics, University College Dublin, Ireland}
\author{Michela Bertolotto}\affil{School of Computer Science and Informatics, University College Dublin, Ireland}
\author{David C. Wilson}\affil{Department of Software and Information Systems, University of North Carolina, USA}

\maketitle

%
 

\keywords{Semantic similarity ensemble; SSE; Lexical similarity; Semantic similarity; Ensemble modelling; Geo-semantics; Expert disagreement; WordNet}


\begin{abstract}
\glsresetall

Computational measures of semantic similarity between geographic terms provide valuable support across geographic information retrieval, data mining, and information integration.
To date, a wide variety of approaches to geo-semantic similarity have been devised. 
A judgement of similarity is not intrinsically right or wrong, but obtains a certain degree of cognitive plausibility, depending on how closely it mimics human behaviour.
Thus selecting the most appropriate measure for a specific task is a significant challenge.
To address this issue, we make an analogy between computational similarity measures and soliciting domain expert opinions, which incorporate a subjective set of beliefs, perceptions, hypotheses, and epistemic biases.
Following this analogy, we define the \emph{\gls{t}} as a composition of different similarity measures, acting as a panel of experts having to reach a decision on the semantic similarity of a set of geographic terms.
The approach is evaluated in comparison to human judgements, and
results indicate that an \gls{t} performs better than the average of its parts.
Although the best member tends to outperform the \gls{shortt}, all \glspl{shortt} outperform the average performance of each \gls{shortt}'s member.
Hence, in contexts where the best measure is unknown, the ensemble provides a more cognitively plausible approach.
\end{abstract}

\section{Introduction}
\label{sec:intro}
\glsresetall
The importance of semantic similarity in \gls{giscience} is widely acknowledged \cite{kuhn:2013:cognling}. 
As diverse information communities generate increasingly large and complex geo-datasets, semantics play an essential role to constrain the meaning of the terms being defined.
The automatic assessment of the semantic similarity of terms, such as \emph{river} and \emph{stream}, enables practical applications in data mining, \gls{gir}, and information integration.
Research in \gls{nlp} and computational linguistics has produced a wide variety of approaches, classifiable as knowledge-based (structural similarity is computed in expert-authored ontologies), corpus-based (similarity is extracted from statistical patterns in large text corpora), or hybrid (combining knowledge and corpus-based approaches) \cite{pedersen:2004:wordnetsim,resnik:1995:using}.
Several similarity techniques have been tailored specifically to geographic information \cite{schwering:2008:approaches}.
In general, a judgement on semantic similarity is not simply right or wrong, but rather shows a certain degree of cognitive plausibility, i.e. a correlation with human behaviour.
Hence, selecting the most appropriate measure for a specific task is non-trivial, and represents in itself a challenge.

From this perspective, a semantic similarity measure bears resemblance with a human expert being summoned to give her opinion on a complex semantic problem.
In domains such as medicine and economic policy, critical choices have be made in uncertain, complex scenarios.
However, disagreement among experts occurs very often, and equally credible and trustworthy experts can hold divergent opinions about a given problem \cite{morgan:1992:uncertainty}.
To overcome decisional deadlocks, an effective solution consists of combining diverse opinions into a representative average.
Instead of identifying a supposedly `best' expert in a domain, an opinion is gathered from a panel of experts, extracting a representative average from their diverging opinions \cite{budescu:2000:confidence}.
Similarly, complex computational problems in machine learning are often tackled with \emph{ensemble methods}, which achieve higher accuracy by combining of heterogeneous models, regressors, or classifiers \cite{rokach:2010:ensembleclassif}.
This idea was first explored in our previous work under the analogy of the \emph{similarity jury} \cite{Ballatore:2012:jury}.

Rather than developing a new measure for geo-semantic similarity, we explore the idea of combining existing measures into a \emph{\gls{t}}.
In order to gain insight about the merits and limitations of the \gls{t}, we conducted a large empirical evaluation, selecting ten WordNet-based similarity measures as a case study.
The ten measures were combined into all of the possible \nensembles{} \glspl{shortt}, exploring the entire combinatorial space.
To measure the cognitive plausibility of each measure and \gls{shortt}, a set of 50 geographic term pairs including 97 unique terms, selected from \gls{osm} and ranked by 203 human subjects, was adopted as ground truth.
The results of this evaluation confirms that, in absence of knowledge about the performance of the similarity measures, the \gls{shortt} approach tends to provide more cognitively plausible results than any individual measure.

The remainder of this paper is organised as follows.
Section \ref{sec:relwork} reviews relevant related work in the areas of geo-semantic similarity and ensemble methods.
Section \ref{sec:wncasestudy} describes the WordNet-based similarity measures selected as a case study.
The \gls{t} is defined in Section \ref{sec:simjury}, while Section \ref{sec:evaluation} presents and discusses the empirical evaluation.
Finally, Section \ref{sec:conclusions} draws conclusions about the \gls{t}, and indicates directions for future work.   

\section{Related work}
\label{sec:relwork}

The ability to assess similarity between stimuli is considered a central characteristic of human psychology.
Hence, it should not come as a surprise that semantic similarity is widely studied in psychology, cognitive science, and \gls{nlp}.
Over the past ten years, a scientific literature on the semantic similarity has emerged in the context of \gls{giscience}  \cite{janowicz:2011:semantics,Ballatore:2012:holistic,Ballatore:2012:geographic}.
Schwering \cite{schwering:2008:approaches} surveyed and classified semantic similarity techniques for geographic terms, including network-based, set-theo\-reti\-cal, and geometric approaches.
Notably, Rodr{\'\i}guez and Egenhofer \cite{rodriguez:2004:comparing} have developed the \gls{mdsmsim} by extending Tversky's set-the\-oretical similarity for geographic terms.
In the area of the Semantic Web, SIM-DL is a semantic similarity measure for spatial terms expressed in \gls{dl} \cite{janowicz:2007:algorithm}.
As these measures are tailored to specific formalisms and data, we selected \gls{wn}-based measures as a more generic case study (see Section \ref{sec:wncasestudy}).


A key element in this article is the combination of different semantic similarity measures, relying on the analogy between computable measures and domain experts.
The idea of combining divergent opinions is not new.
Indeed, expert disagreement is not an exceptional state of affairs, but rather the norm in human activities characterised by uncertainty, complexity, and trade-offs between multiple criteria \cite{morgan:1992:uncertainty}.
As Mumpower and Stewart \cite{mumpower:1996:expert} put it, the ``character and fallibilities of the human judgement process itself lead to persistent disagreements even among competent, honest, and disinterested experts'' (p. 191).
From a psychological perspective, in cases of high uncertainty and risk (e.g. choosing medical treatments and long term investments), decision makers consult multiple experts, and try to obtain a representative average of divergent expert judgements \cite{budescu:2000:confidence}.
In the context of risk analysis, mathematical and behavioural models have been devised to elicit judgements from experts, suggesting that simple mathematical methods such as the average perform quite well \cite{clemen:1999:combining}.
The underlying intuition has been controversially labelled as `wisdom of crowds,' and can account for the success of some crowdsourcing applications \cite{surowiecki:2005:wisdom}.

In complex domains such as econometrics, genetics, and meteorology, \emph{ensemble methods} aggregate different models of the same phenomenon, trying to overcome the limitations of each model.
In the context of machine learning, a wide variety of ensemble methods have been devised and evaluated \cite{rokach:2010:ensembleclassif}.
Such methods aim at generating a single classifier from a set of classifiers applied to the same problem, maximising its overall accuracy and robustness \cite{opitz:1999:popularensemble}.
Similarly, clustering ensembles obtain a single partitioning of a set of objects by aggregating several partitionings returned by different clustering techniques \cite{strehl:2003:clusterensembles}.
In computational biology, ensemble approaches are currently being used to compute the similarity of proteins \cite{keiser:2007:relatingprotein}.

Forecasting complex phenomena can also benefit from ensemble methods.
Armstrong \cite{armstrong:2001:combiningforecasts} pointed out that ``combining forecasts is especially useful when you are uncertain about the situation, uncertain about which method is most accurate, and when you want to avoid large errors'' (p. 417). 
Notably, a study of the Blue Chip Economic Indicators survey indicates that forecasts issued by a panel of seventy economists tended to outperform all the seventy individual forecasts \cite{bauer:2003:forecast}.
To date, we are not aware of studies that explores systematically the possibility of combining semantic similarity measures through an ensemble method.
The next section describes in detail the similarity measures that we selected as a case study.

\section{WordNet similarity measures}
\label{sec:wncasestudy}
\glsresetall

In this study, we selected \gls{wn}-based semantic similarity measures as a case study for our ensemble technique, the \gls{t}.
In the context of \gls{nlp}, \gls{wn} \cite{fellbaum:2010:wordnetsurvey} is a well-known knowledge base for the computation of semantic similarity.
Numerous knowledge-based approaches exploit its deep taxonomic structure for nouns and verbs \cite{leacock:1998:combining,resnik:1995:using,lin:1998:information,wu:1994:verbs,banerjee:2002:adapted}.
From a geo-semantic viewpoint, \gls{wn} terms have been mapped to \gls{osm} \cite{Ballatore:2013:groundinglod}.
Table \ref{table:relwork_WordNetSimilarities} summarises the salient characteristics of ten popular \gls{wn}-based measures.
In order to compute the similarity scores, each measure adopts a different strategy.
Seven measures relies on the \emph{shortest path} between terms in the noun/verb taxonomy, assuming that the number of edges is inversely proportional to the similarity of terms.
This approach is limited by the variability in the path lengths in the different semantic areas of \gls{wn}, determined by arbitrary choices and biases of the knowledge base's owners.
Paths in dense, well-developed parts of the taxonomy tend to be longer than those in shallow, sparse areas, making the direct comparison of term pairs from different areas problematic.
Missing edges between terms make the score drop to $0$.

To overcome these limitations, three measures include the \emph{information content} of the two terms and that of the \emph{least-common subsumer}, i.e. the more specific term that is an ancestor to both target terms \cite[e.g.][]{resnik:1995:using}.
Hence, at the same path length, terms with a very specific subsumer (`building') are considered to be more similar than terms with a generic subsumer (`thing').
Although this approach mitigates the issues of the shortest paths, a new issue lies in the extraction of the information content from a text corpus.
Text corpora tend to be biased towards specific semantic fields, underestimating the specificity of terms contained in those fields, resulting in skewed similarity scores.
An alternative approach that do not rely on taxonomy paths consists of comparing the term \emph{glosses}, i.e. the lexical definition of terms.
Definitions can be compared in terms of word overlap (terms that are defined with the same words tend to be similar), or with co-occurrence patterns in a text corpus (terms that are defined with co-occurring words tend to be similar) \cite{patwardhan:2006:using}.
The results of this approach are sensitive to noise in the definitions (e.g. very frequent or rare words that skew the scores), and to the arbitrary nature of definitions, which can under- or over-specified.

Empirical research suggests that the performance of these measures largely depends on the specific ground-truth dataset utilised in the evaluation \cite{mihalcea:2006:corpus}.
Therefore, these measures constitute a striking example of alternative models of the same phenomenon, none of which can be considered to be uncontroversially better than the others.
Each measure is sensitive to specific biases in the knowledge base, and tends to reflect these biases in the similarity scores.
For this reason, we consider these measures to be a suitable case study for the ensemble approach, formally defined in the next section. 


\begin{table}[t]
\begin{tabular}{c|P{9em}P{11em}ccc}
    \hline
	Name & Reference & Description & SPath & Gloss & InfoC \\ \hline\hline
	path & Rada \emph{et al.} \cite{rada:1989:development} & Edge count in the semantic network & \tick & &  \\\hline
	lch & Leacock and Chodorow \cite{leacock:1998:combining} & Edge count scaled by depth & \tick &  &  \\\hline
	res & Resnik \cite{resnik:1995:using} & Information content of $lcs$ & \tick &  & \tick  \\\hline
	jcn & Jiang and Conrath \cite{jiang:1997:semantic} & Information content of $lcs$ and terms & \tick &  & \tick \\\hline
	lin & Lin \cite{lin:1998:information} & Ratio of information content of $lcs$ and terms & \tick &  & \tick  \\\hline
	wup & Wu and Palmer \cite{wu:1994:verbs} & Edge count between $lcs$ and terms & \tick & &  \\\hline
	hso & Hirst and St-Onge \cite{hirst:1998:lexical} & Paths in lexical chains & \tick &  &  \\\hline
	lesk & Banerjee and Pedersen \cite{banerjee:2002:adapted} & Extended gloss overlap &  & \tick &  \\\hline
	~vector~ & Patwardhan and Pedersen \cite{patwardhan:2006:using} &  Second order co-occurrence vectors &  & \tick &  \\ \hline 
	vectorp  & Patwardhan and Pedersen \cite{patwardhan:2006:using} & Pairwise second order co-occurrence vectors  &  & \tick & \\\hline 
\end{tabular} 
\caption[WordNet-based similarity measures]{WordNet-based similarity measures. \emph{SPath}: shortest path; \emph{Gloss}: lexical definitions (glosses); \emph{InfoC}: information content; \emph{lcs}: least common subsumer.}
\label{table:relwork_WordNetSimilarities}
\end{table}

\glsresetall
\section{The \protect\gls{t}}
\label{sec:simjury}
\glsresetall
A computable measure of semantic similarity can be seen as a human domain expert summoned to rank pairs of terms, according to her subjective set of beliefs, perceptions, hypotheses, and epistemic biases.
When the performance of an expert can be compared against a gold standard, it is a reasonable policy to trust the expert showing the best performance.
Unfortunately, such gold standards are difficult to construct and validate, and the choice of most appropriate expert remains highly problematic in many contexts.
To overcome this issue, we propose the \emph{\gls{t}}, a technique to combine different semantic similarity measures on the same set of terms.
This \gls{shortt} of measures can be intuitively seen as a jury or a panel of human experts deliberating on a complex case \cite{Ballatore:2012:jury}.
Formally, the similarity function $sim$ quantifies the semantic similarity of a pair of geographic terms $t_a$ and $t_b$ ($sim(t_a,t_b) \in [0,1]$).
Set $P$ contains all term pairs whose similarity needs to be assessed, while set $M$ contains a set of selected semantic similarity measures from which the \glspl{shortt} will be formed:
\begin{eqnarray}
P = \{ \langle t_{a1} t_{b1} \rangle, \langle t_{a2} t_{b2} \rangle ~\ldots ~ \langle t_{an} t_{bn} \rangle \} \\
M = \{ sim_{1}, sim_{2} ~\ldots ~ sim_{m} \} \nonumber
\end{eqnarray}

\noindent A measure $sim$ from $M$ applied to $P$ maps the set of pairs to a set of scores $S_{sc}$, which can then be converted into rankings $S_{rk}$, from the most similar (e.g. \emph{stream} and \emph{river}) to the least similar (e.g. \emph{stream} and \emph{restaurant}):
\begin{eqnarray}
sim(P) \rightarrow S_{sc} = \{ s_1, s_2 \ldots s_n \} ~~~~ s \in \mathbb{R}_{\geq 0}\\
rank(S_{sc}) \rightarrow S_{rk} =  \{ r_1, r_2 \ldots r_n \}\nonumber 
\end{eqnarray}

\noindent For example, a measure $sim \in M$ applied to a set of three pairs $P$ might return $S_{sc} = \{ .45, .13, .91 \}$, corresponding to rankings $S_{rk} = \{ 2, 3, 1 \}$.
The rankings $S_{rk}(P)$ can be used to assess the cognitive plausibility of $sim$ against a human-generated rankings $H(P)$.
The cognitive plausibility of $sim$ can be estimated with the Spearman's correlation $\rho \in [-1,1]$ between $S_{rk}(P)$ and $H_{rk}(P)$.
If $\rho$ is close to 1 or -1, $sim$ is highly plausible, while if $\rho$ is close to 0, $sim$ shows no correlation with human behaviour.

\glsreset{t}
In this context, a \emph{\gls{t}} is defined as a set $E$ of unique semantic similarity measures:
\begin{eqnarray}
E = \{ sim_{1}, sim_{2} ~\ldots ~ sim_{k} \},~~ \forall j \in \{1,2 \ldots k\}: sim_j \in M\\
\forall i \in \{1,2 \ldots |M|-1\}: ~ sim_i \ne sim_{i+1},~~~~~~~ k\leq m, |E| \leq |M| \nonumber
\end{eqnarray}

\noindent For example, considering the ten measures in Table \ref{table:relwork_WordNetSimilarities}, \gls{shortt} $E_a$ has two members $\{ jcn, lesk \}$, while \gls{shortt} $E_b$ has three members $\{ jcn, res, wup \}$. 

Several techniques have been discussed to aggregate rankings, using either unsupervised or supervised methods.
Clemen and Winkler \cite{clemen:1999:combining} stated that simple mathematical methods, such as the average, tend to perform quite well to combine expert judgements in risk assessment.
Hence, we define two aggregation approaches $A$ to compute the rankings of \gls{shortt} $E$:
\begin{enumerate}
  \item Mean of the similarity scores: $A_{s} = rank( mean(S_{sc1}, S_{sc2} \ldots S_{scn}))$ 
  \item Mean of the similarity rankings: $A_{r} = rank( mean(S_{rk1}, S_{rk2} \ldots S_{rkn} ) )$ 
\end{enumerate}


\noindent The first approach, $A_{s}$, combines directly the similarity scores, while the second approach flattens the scores into equidistant rankings.
Rankings contain less information than scores: for example, scores $\{.01,.02,.98,.99\}$ and $\{.51,.52,.53,.54\}$ have very different distributions, but result in the same rankings $\{1,2,3,4\}$.
For this reason, in some cases, $A_s \ne A_r$.
If two measures on five term pairs generate the scores $S_{sc1} = \{.9,.9,.38,.44,.31\}$ and $S_{sc2} = \{.28,.47,.14,.61,.36\}$, the resulting $A_s$ is $\{4,5,1,3,2\}$, whilst $A_r$ is $\{3,5,1,4,2\}$.

A given similarity measure has a cognitive plausibility, i.e. the ability to approximate human judgement.
A traditional approach to quantify the cognitive plausibility of a measure consists of comparing rankings against a human-generated ground truth \cite{ferrara:2013:evalrelatedness}.
The ranked similarity scores are compared with the rankings or ratings returned by human subjects on the same set of term pairs.
Following this approach, we define $\rho_{sim}$ as the correlation of an individual measure $sim$ (i.e. an \gls{shortt} of size one) with human-generated rankings $H_{rk}$, while $\rho_{E}$ the correlation of the judgement obtained from an \gls{shortt} $E$.
When knowledge of $\rho_{sim}$ is available for the current task, the optimal $sim \in M$ can be simply the $sim$ having highest $\rho_{sim}$.
However, in real settings this knowledge is often absent, or incomplete, or unreliable.
The same semantic similarity measure can obtain considerably different degrees of cognitive plausibility based on the specific dataset in consideration.
In such contexts of limited information, the \gls{t} offers a viable alternative to an arbitrary selection of a $sim$ from $M$. 
The empirical evidence discussed in the next section supports this claim.  

\section{Evaluation}
\label{sec:evaluation}

This section discusses an empirical evaluation conducted on the \gls{t} in real settings.
The purpose of this evaluation is to assess the performance of the \gls{t} in detail, highlighting strengths and weaknesses.
Ten semantic similarity measures are tested on a set of pairs of geographic terms utilised in \gls{osm}.  
A preliminary evaluation of an analogous technique on a small scale was conducted in \cite{Ballatore:2012:jury}.
Ensembles of cardinalities 2,3, and 4 were generated from eight similarity measures, for a total of 154 \glspl{shortt}.
The evaluation described below is conducted on a larger scale, adopting a larger set of geographic terms, ranked by 203 human subjects as ground truth.
To obtain a complete picture of \gls{shortt}'s performance, the entire combinatorial space is considered, for a total of \nensembles{} unique \glspl{shortt}.
The remainder of this section outlines the evaluation criteria by which the performance of the \gls{t} is assessed (Section \ref{sec:evalcriteria}), the human-generated ground truth (Section \ref{sec:groundtruth}), the experiment set-up (Section \ref{sec:expsetup}), and the empirical results obtained, including a comparison with the preliminary evaluation (Section \ref{sec:expresults}).

\subsection{Evaluation criteria}
\label{sec:evalcriteria}

The performance of an ensemble $E$ is measured on its cognitive plausibility $\rho_{E}$, with respect to the plausibility of its individual members $\rho_{sim}$. 
Intuitively, an \gls{shortt} succeeds when it provides rankings that are more cognitively plausible than those of its members.
Four criteria are formally defined in this evaluation:

\begin{itemize}
  \item \textbf{Total success.} The plausibility of the \gls{shortt} is strictly greater than all of its members: $\forall sim \in E: \rho_{E} > \rho_{sim}$
  \item \textbf{Partial success.} The plausibility of the \gls{shortt} is strictly greater than a member: $\exists sim \in E: \rho_{E} > \rho_{sim}$
  \item \textbf{Success over mean.} The plausibility of the \gls{shortt} is strictly greater than the mean plausibility of its members: $\rho_{E} > mean(\rho_{sim_1},\rho_{sim_2}\ldots\rho_{sim_n})$
  \item \textbf{Success over median.} The plausibility of the \gls{shortt} is strictly greater than the median plausibility of its members: $\rho_{E} > median(\rho_{sim_1},\rho_{sim_2}\ldots\rho_{sim_n})$
\end{itemize}

\subsection{Ground truth}
\label{sec:groundtruth}

In order to assess the cognitive plausibility of the similarity measures and the \glspl{shortt}, a human-generated ground truth has to be selected.
In the preliminary evaluation described, a human-generated set of similarity rankings was extracted from an existing dataset \cite{Ballatore:2012:jury}.
That dataset contains similarity rankings of 50 term pairs, over on 29 geographic terms, originally collected by Rodr{\'\i}guez and Egenhofer \cite{rodriguez:2004:comparing}, and is available online.\footurl{http://github.com/ucd-spatial/Datasets}
In order to provide a thorough assessment of the \gls{t} in the present article, a new and larger human-generated dataset was adopted as ground truth.

As part of a wider study on geo-semantic similarity, we selected 50 pairs of geographic terms commonly used in \gls{osm}, including 97 man-made and natural features.
The terms were subsequently mapped to the corresponding terms in \gls{wn}, as exemplified in Table \ref{table:geresid_terms}. 
A Web-based survey was subsequently prepared on the set of 50 term pairs, asking human subjects to rate the pairs' similarity on a five-point Likert scale, from \emph{very dissimilar} to \emph{very similar}.
In order to be understandable by any native speaker of English, regardless of knowledge of the geographic domain, the survey only included common and non-technical terms, aiming to collect a generic set of geo-semantic judgements.
The survey was disseminated online through mailing lists, and obtained valid responses from 203 human subjects.
The subjects' ratings for each pair were normalised on a $[0,1]$ interval and averaged, obtaining human-generated similarity scores $H_{sc}$, then ranked as $H_{rk}$.
Table \ref{table:geresid_pairs} outlines a sample of term pairs, with the similarity score and ranking assigned by the 203 human subjects.
This dataset was utilised as ground truth in the experiment outlined in the next section.

\begin{table}[t] 
\footnotesize
\begin{tabular}{lll}
\hline
Term & \gls{osm} tag $~~$ & \gls{wn} synset \\
\hline
 bay & natural=bay & bay\#n\#1\\ 
 canal & waterway=canal & canal\#n\#3\\ 
 city & place=city & city\#n\#1\\ 
 post box & amenity=post\_box & postbox\#n\#1\\ 
floodplain & natural=floodplain & floodplain\#n\#1\\ 
historic castle & historic=castle & castle\#n\#2\\ 
motel & tourism=motel & motel\#n\#1\\ 
supermarket & shop=supermarket & supermarket\#n\#1\\ 
\ldots & \ldots & \ldots \\
\hline
 \end{tabular} 
\caption{Sample of the 97 terms extracted from \gls{osm} and mapped to \gls{wn}.}
\label{table:geresid_terms}
\end{table}

\begin{table}[t] 
\begin{tabular}{llcc}
\hline
 Term A & Term B &  $~~H_{sc}~~$  &  $~~H_{rk}~~$ \\
\hline
  motel & hotel &  .90 & 1 \\ 
 public transport station $~~$ & railway platform &  .81 & 2\\ 
 stadium & athletics track &  .76 & 3\\ 
 theatre & cinema &  .87 & 4\\ 
 art shop & art gallery &  .75 & 5\\ 
 \ldots & \ldots & \ldots & \ldots \\
water ski facility & office furniture shop &  .05 & 46\\
greengrocer & aqueduct &  .03 & 47\\ 
interior decoration shop & tomb  & .05 & 48\\ 
 political boundary & women's clothes shop  & .02 &49\\ 
 nursing home & continent &  .02 & 50\\
\hline 
 \end{tabular} 
\caption{Human-generated similarity scores ($H_{sc}$) and rankings ($H_{rk}$) on 50 term pairs}
\label{table:geresid_pairs}
\end{table}  

\subsection{Experiment setup}
\label{sec:expsetup}

To explore the performance of an \gls{t} versus individual measures, we selected a set of ten WordNet-based similarity measures as a case study.
Table \ref{table:expsetup} summarises the resources involved in this experiment. 
The ten similarity measures were not applied directly to the term pairs, but they were applied to the their lexical definitions, using a paraphrase-detection technique \cite{Ballatore:2013:lexicaldefsim}.\footnote{The \emph{WordNet::Similarity} tool \cite{pedersen:2004:wordnetsim} was used to compute the similarity scores.}
\begin{table}[t]
 \begin{tabular}{rcl} 
 \hline
 10 similarity measures $sim \in M$: & $~~$ & $\{ jcn, lch, hso, lesk, lin, path,~~~~~~~~~~~~~~$ \\
   && $~~res, vector, vectorp, wup \}$ (see Table \ref{table:relwork_WordNetSimilarities})\\ \hline
 9 \gls{shortt} cardinalities $|E|$:  && $\{2,3,4,5,6,7,8,9,10\}$ \\ \hline
	Number of unique \glspl{shortt} $E$:  && \{$45,120,$ $210,$ $252,$ $210,$ $120,45,9,1$\}; Total: \nensembles{} \\\hline
	2 types of \glspl{shortt} $E$:  && \gls{shortt} of scores $E_s$ and \gls{shortt} of rankings $E_r$ \\\hline
	Ground truth: & & 50 term pairs ranked by \\
	 && 203 human subjects by semantic similarity\\\hline
	4 evaluation criteria: & & (a) total success; (b) partial success;\\
	 && (c) success over mean; (d) success over median\\
	\hline
  \end{tabular} 
 \caption{Experiment setup}
 \label{table:expsetup}
 \end{table}
In order to explore the space of all the possible \glspl{shortt}, we considered the entire range of \gls{shortt} sizes $|E| \in \{2,3 \ldots 10\}$ for $M$.
The entire power set of $M$ was computed.
Increasing the \gls{shortt} cardinality from 2 to 10, respectively $45,120,$ $210,$ $252,$ $210,$ $120,45,9,1$ \glspl{shortt} were generated, for a total \nensembles{} \glspl{shortt}.
The experiment was carried out through the following steps:
\begin{enumerate}
  \item Compute $S_{sc}$ and $S_{rk}$ for each of the ten measures on the 50 term pairs from \gls{osm}. 
  \item Generate \nensembles{} \glspl{shortt}, combining the measures on either similarity scores ($E_s$) or rankings ($E_r$).
  \item For each of the ten measures, compute the cognitive plausibility $\rho_{sim}$ against human-generated rankings $H_{rk}$.
  \item For each of the \nensembles{} \glspl{shortt}, compute the cognitive plausibility $\rho_E$ against $H_{rk}$.
  \item Compute the four evaluation criteria (total success, partial success, success over mean, success over median) for each measure and \gls{shortt}.
\end{enumerate}


\begin{table}[p]
\scriptsize
\begin{tabular}{P{5.3em}|r|rrrrrrrrrr|r}
\hline 
 & $|E|$ & vector & lch & path & hso & wup & vecp & res & lesk & jcn & lin & mean\\
 \hline
$\rho_{sim}$ & $-$ & .737 & .727 & .727 & .708 & .663 & .641 & .635 & .628 & .588 & .562 & .662\\ 
\hline
Total & 2 & 33.3 & 22.2 & 22.2 & 11.1 & 22.2 & 33.3 & 22.2 & 55.6 & 11.1 & 11.1 & 24.4\\ 
success  & 3 & 27.8 & 22.2 & 27.8 & 27.8 & 36.1 & 19.4 & 36.1 & 41.7 & 11.1 & 8.3 & 25.8\\ 
(\%) & 4 & 11.9 & 17.9 & 16.7 & 17.9 & 22.6 & 10.7 & 23.8 & 20.2 & 6.0 & 4.8 & 15.2\\ 
 & 5 & 8.7 & 11.1 & 13.5 & 11.9 & 12.7 & 7.1 & 16.7 & 12.7 & 2.4 & 2.4 & 9.9\\ 
 & 6 & 8.7 & 8.7 & 7.9 & 9.5 & 6.3 & 4.0 & 9.5 & 6.3 & 0.0 & 0.8 & 6.2\\ 
 & 7 & 3.6 & 4.8 & 4.8 & 4.8 & 3.6 & 3.6 & 4.8 & 3.6 & 0.0 & 0.0 & 3.3\\ 
 & 8 & 2.8 & 2.8 & 2.8 & 2.8 & 2.8 & 0.0 & 2.8 & 2.8 & 0.0 & 2.8 & 2.2\\ 
 & 9 & 0.0 & 0.0 & 0.0 & 0.0 & 0.0 & 0.0 & 0.0 & 0.0 & 0.0 & 0.0 & 0.0\\ 
 & 10 & 0.0 & 0.0 & 0.0 & 0.0 & 0.0 & 0.0 & 0.0 & 0.0 & 0.0 & 0.0 & 0.0\\ 
 \hdashline
Mean & All & 10.8 & 10.0 & 10.6 & 9.5 & 11.8 & 8.7 & 12.9 & 15.9 & 3.4 & 3.4 & 9.7\\ 
\hline 
Partial  & 2 & 33.3 & 22.2 & 22.2 & 33.3 & 66.7 & 77.8 & 77.8 & 100.0 & 88.9 & 100.0 & 62.2\\ 
success & 3 & 27.8 & 25.0 & 30.6 & 58.3 & 86.1 & 94.4 & 97.2 & 97.2 & 100.0 & 100.0 & 71.7\\ 
 (\%) & 4 & 11.9 & 26.2 & 23.8 & 50.0 & 95.2 & 97.6 & 98.8 & 100.0 & 100.0 & 100.0 & 70.3\\ 
 & 5 & 8.7 & 19.8 & 22.2 & 58.7 & 95.2 & 100.0 & 100.0 & 100.0 & 100.0 & 100.0 & 70.5\\ 
 & 6 & 8.7 & 18.3 & 17.5 & 56.3 & 98.4 & 100.0 & 100.0 & 100.0 & 100.0 & 100.0 & 69.9\\ 
 & 7 & 3.6 & 19.0 & 19.0 & 67.9 & 100.0 & 100.0 & 100.0 & 100.0 & 100.0 & 100.0 & 71.0\\ 
 & 8 & 2.8 & 11.1 & 11.1 & 63.9 & 100.0 & 100.0 & 100.0 & 100.0 & 100.0 & 100.0 & 68.9\\ 
 & 9 & 0.0 & 22.2 & 22.2 & 55.6 & 100.0 & 100.0 & 100.0 & 100.0 & 100.0 & 100.0 & 70.0\\ 
 & 10 & 0.0 & 0.0 & 0.0 & 100.0 & 100.0 & 100.0 & 100.0 & 100.0 & 100.0 & 100.0 & 70.0\\
 \hdashline 
Mean & All & 10.8 & 18.2 & 18.7 & 60.4 & 93.5 & 96.6 & 97.1 & 99.7 & 98.8 & 100.0 & 69.4\\ 
\hline
Succ. mean & All & 100.0 & 100.0 & 100.0 & 100.0 & 100.0 & 100.0 & 100.0 & 100.0 & 100.0 & 100.0 & 100.0\\ 
\hline
Succ. med. & All & 96.6 & 95.6 & 96.2 & 97.9 & 99.7 & 98.9 & 99.9 & 99.7 & 97.8 & 97.1 & 98.0\\
\hline
 \end{tabular} 
\caption{Overall results of the experiment, including cognitive plausibility $\rho_{sim}$, and the four evaluation criteria. 
\emph{Succ. mean}: success over mean; \emph{Succ. med.}: success over median.}
\label{table:eval_results}
\end{table}

\begin{figure}[p]
  \centering
  \includegraphics[width=35em]{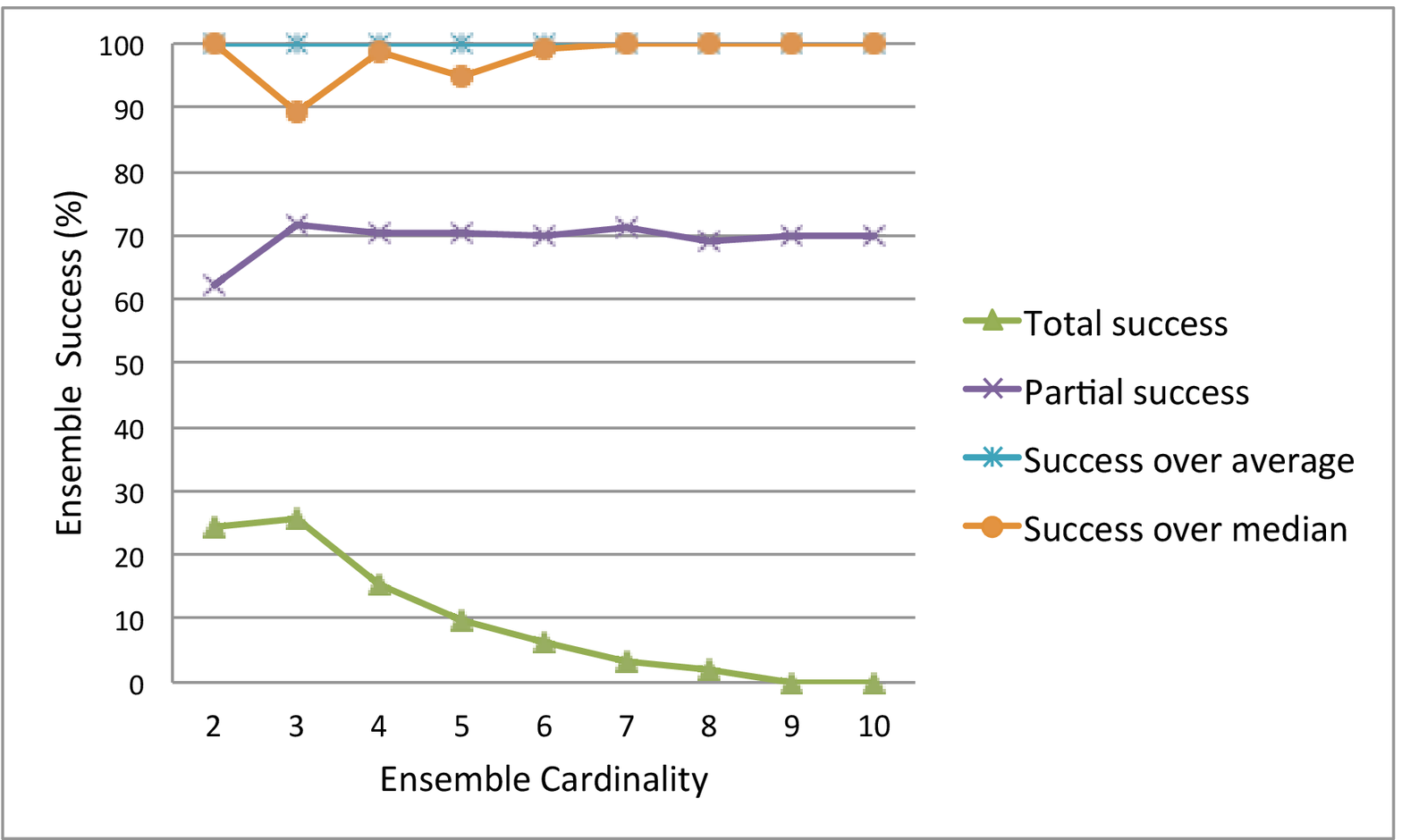}
  \caption{The four evaluation criteria w.r.t. \gls{shortt} cardinality}
  \label{figure:success_trends}
\end{figure}

\subsection{Experiment results}
\label{sec:expresults}


The experiment was carried out on two types of \gls{shortt}, once with $A_s$ (mean of scores), and once with $A_r$ (mean of rankings).
These two approaches obtained very close results, with a slightly better performance for $A_r$, with each evaluation criterion always within a $5\%$ distance from $A_s$.
To avoid repetition, only cases with $A_r$ are included in the discussion.
All the cognitive plausibility correlations obtained statistically significant results at $p < .01$. 
The experiment results are summarised in Table \ref{table:eval_results}, showing the cognitive plausibility of each measure, and the four evaluation criteria across all the \gls{shortt} cardinalities.
For example, the \glspl{shortt} of cardinality $2$ containing measure $wup$ obtains partial success in $86.1\%$ of the cases.
The cognitive plausibility of the ten measures are in range $\rho \in [.562,.737]$, where $vector$ is the best measure, and $lin$ the worst.
Whilst total and partial success change considerably and are fully reported, the success over mean and median obtain homogenous results and only the means are included in the table. 
The general trends followed by the evaluation criteria are depicted in Figure \ref{figure:success_trends}.

\begin{figure}[t]
  \centering
  \includegraphics[width=35em]{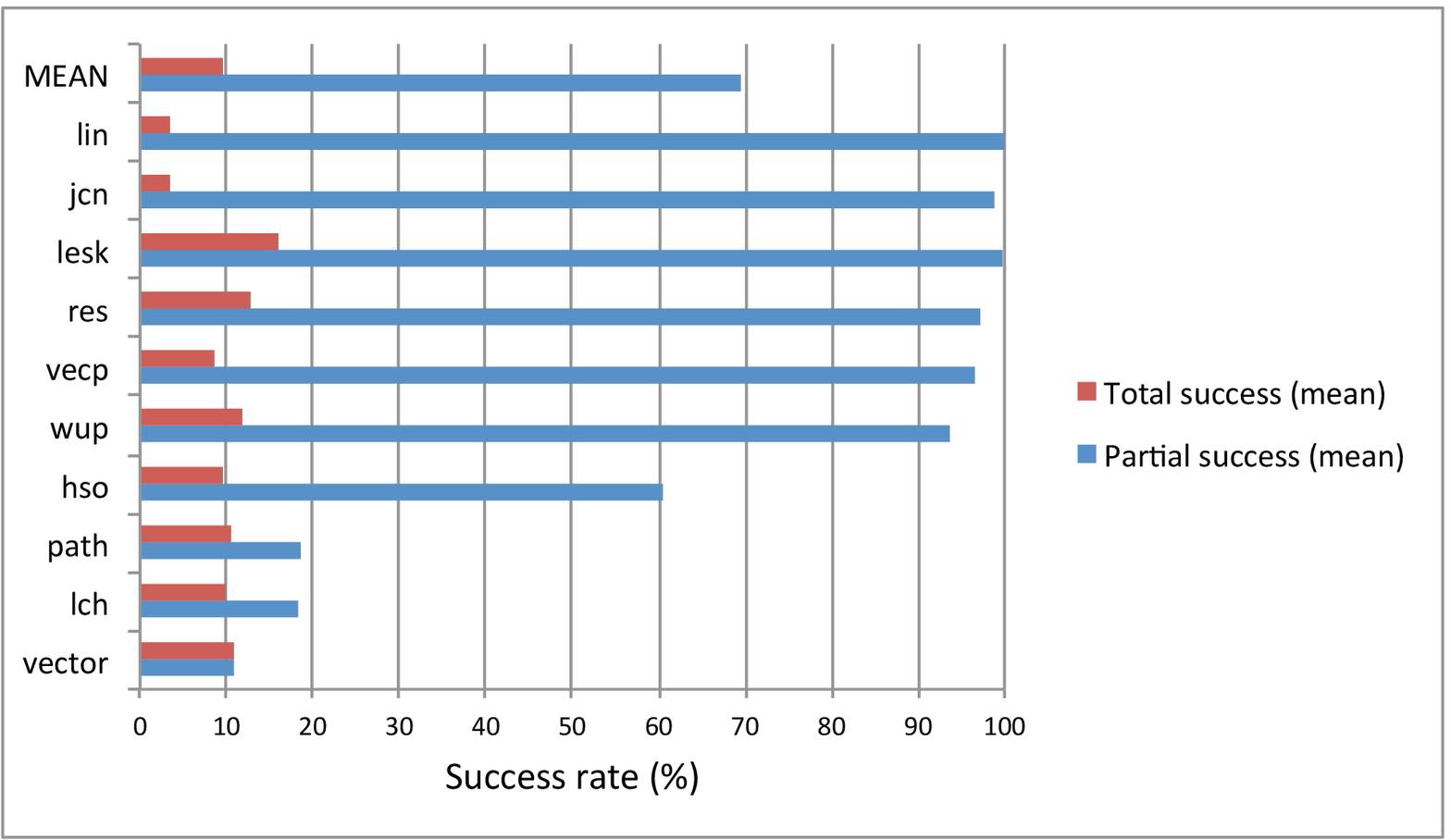}
  \caption{\Gls{shortt} total and partial success w.r.t. similarity measures $sim$}
  \label{figure:measures_success}
\end{figure} 

\paragraph{Total success.}
The total success for the \nensembles{} \glspl{shortt} falls in interval $[0,55.6]$ percent, with a mean of $9.7\%$.
On average, small cardinalities (2 and 3) obtain the best total success rate ($\approx 25\%$).
As the cardinality increases, the total success decreases rapidly, dropping below $10\%$ with cardinality greater than 4.
This makes sense intuitively, as the larger the \gls{shortt}, the less likely the \gls{shortt} can outperform every single member.
The total success varies across the different measures too, falling in interval $[3.4,15.9]$.
No statistically significant correlation exists between a measure's cognitive plausibility and its rate of total success.
In other words, \glspl{shortt} containing the best measures do not necessarily have better or worse total success rate. 
Although \glspl{shortt} do not tend to outperform all of their members, the plausibility of an \gls{shortt} is never lower than that of all of its members, $~\exists sim \in E : \rho_{E} > \rho_{sim}$.

\paragraph{Partial success.}
Partial success rate is considerably greater than that of total success.
Over the entire space of \glspl{shortt}, the partial success rate varies widely between $0\%$ and $100\%$, with an global average of $\approx 70\%$.
The \glspl{shortt}' cardinality has no clear impact on the mean partial success rate, which remains in the interval $[62.2,71.7]$ both with small and large \glspl{shortt}.
Unlike total success, partial success rate is affected by each measure's cognitive plausibility $\rho_{sim}$.
The top measures in $M$ ($vector$, $lch$, and $path$) obtain low partial success ($< 20\%$), whereas \glspl{shortt} consistently outperform the bottom measures ($100\%$).

The average partial success rates bear strong inverse correlation with the measures' plausibility, i.e. $\rho = -.87 ~ (p < .05)$.
Ensembles tend to outperform the worst measures, and tend to be outperformed by the top measures.
The total and partial success of each measure is displayed in Figure \ref{figure:measures_success}. 
We note that the three top measures do not benefit from being aggregated within the \gls{shortt}, whereas all the others do.
While in this experiment a ground truth is given, in many real-world settings the best measures are unknown, and therefore the \gls{t} constitutes a viable alternative to the arbitrary selection of a measure.
In particular, \glspl{shortt} of cardinality 3 obtain optimal results over other cardinalities.

\begin{figure}[t]
  \centering
  \includegraphics[width=28em]{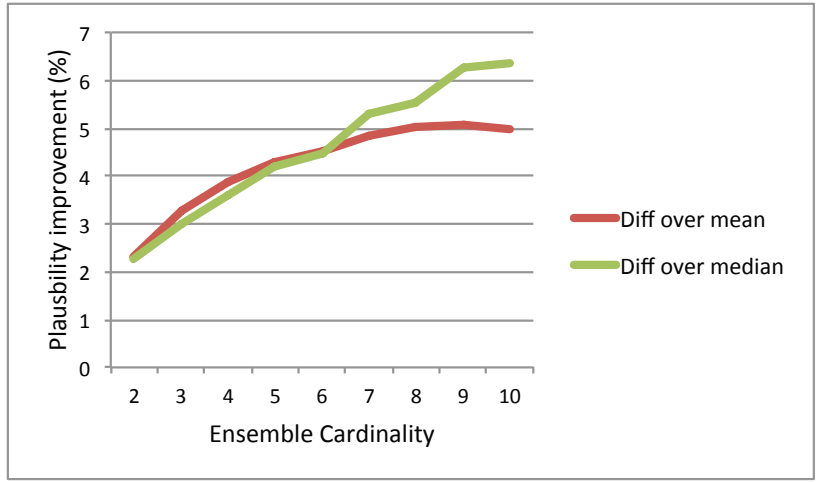}
  \caption{Improvement in cognitive plausibility of the \glspl{shortt} over the mean and median of their members' plausibility}
  \label{figure:plaus_improvement}
\end{figure}

\paragraph{Success over mean and median.}
Unlike total and partial success, the success of \glspl{shortt} over the mean and median of their members' plausibilities is consistent.
All \nensembles{} \glspl{shortt} obtain higher plausibility than the mean of their members' plausibilities ($100\%$).
Similarly, $98\%$ of the \glspl{shortt} are more plausible than the median of their members' plausibilities.
Hence, an \gls{shortt} is more than the mean (or the median) of its parts.
In order to quantify more precisely the advantage of the \glspl{shortt} over the mean of their members' plausibilities, we computed the difference between the \gls{shortt}'s plausibility $\rho_{E}$ and the mean (or median) of all the $\rho_{sim}$, where $sim \in E$.
On average, the \glspl{shortt}' plausibility is $.042$ higher than the mean of their members ($+4.2\%$), and $.046$ over the median ($+4.6\%$).
Figure \ref{figure:plaus_improvement} depicts the advantage of the \gls{shortt} in terms of cognitive plausibility over mean and median, with respect with the cardinality of the \gls{shortt}.
The advantage is directly proportional to the \gls{shortt}'s size, i.e. the larger the \gls{shortt}, the larger the improvement over mean and median.
In other words, by combining the rankings, the \gls{shortt} reduces the weight of individual bias, converging towards a shared judgement.
Such shared judgement is not necessarily the best fit in absolute terms, but tends to be more reliable than most individual judgements.

\paragraph{Comparison with preliminary experiment.}
To further assess the \gls{t}, the empirical evidence described above can be compared with the preliminary evaluation we conducted in \cite{Ballatore:2012:jury}, discussing their commonalities and differences.
That evaluation included only eight of the ten \gls{wn}-based similarity measures, on \glspl{shortt} of cardinality 2, 3, and 4, called \emph{similarity juries}.
These measures and \glspl{shortt} were compared against an existing similarity dataset, originally collected by Rodr{\'\i}guez and Egenhofer \cite{rodriguez:2004:comparing}.
The salient characteristics of the two evaluations are summarised in Table \ref{table:expcomp}.
The comparison of the two evaluations reveals that the same general trends are observable across the board.
The total success of the current evaluation appears to be lower than in the preliminary evaluation, and this is because the current evaluation includes larger \glspl{shortt}, which tend to have lower total success than the small \glspl{shortt} of cardinality smaller than 5.
On average, the partial success rates are very similar in both evaluations ($\approx 70\%$).
The success over mean is very high in both evaluations, consistently falling between from $93\%$ to $100\%$.

Although the mean plausibility of the measures is consistent across the two evaluations, the relative performances of the individual measures vary widely.
Notably, the measure $jcn$ is the most plausible measure in the preliminary evaluation, while being the second-last in the current evaluation.
Similarly, $vector$ is the top measure in the current evaluation, and ranks among the worst in the preliminary evaluation.
By contrast, $lch$, $wup$, and $lesk$ maintain almost the same relative position in terms of cognitive plausibility.
The two sets of plausibilities do not show any statistically significant correlation (Spearman's $\rho \approx .1$).
Although the measures fall within a similar range in both evaluations, it is difficult to identify measures that are always optimal or inadequate.
These results confirm the difficulty of identifying optimal semantic similarity measures, suggesting that the \gls{t} offers a way to proceed in a context of limited and uncertain information.

\begin{table}[t]
\begin{tabular}{rrr} 
\hline
Input and output &$~~$ Preliminary  & Current \\
parameters & evaluation & $~~$ evaluation \\
\hline
Ground truth: geographic terms  & $~~$ 29 & 97 \\
Ground truth: term pairs & $~~$ 50  & $~~$50 \\
Ground truth: human subjects & $~~$ 72 & $~~$ 203 \\
Similarity measures & 8 & 10 \\
Similarity \glspl{shortt} & 154 & \nensembles{} \\
Cardinalities & $\{2,3,4\}$ & $\{2,3\ldots10\}$ \\
\hdashline
Measures' plausibility (mean $\rho$) & $.62$ & $.66$ \\
Measures' plausibility (range $\rho$) & $[.45,.72]$ & $[.56,.74]$ \\
Total success (range $\%$)   & $[28.6,46.1]$ & $[0,55.6]$ \\
Total success (mean $\%$) 	& $34.8$ & $9.7$ \\
Partial success (range $\%$) & $[55,87.2]$ & $[0,100]$ \\
Partial success (mean $\%$) 	& $73.3$ & $69.4$ \\
Success over mean (mean $\%$) & $93.2$ & $100$ \\

\hline
  \end{tabular} 
 \caption{Comparison between the preliminary evaluation in \cite{Ballatore:2012:jury} and the evaluation in this article.}
 \label{table:expcomp}
 \end{table}

\section{Conclusions}
\label{sec:conclusions}
\glsresetall
In this paper we have outlined, formalised, and evaluated the \emph{\gls{t}}, a combination technique for semantic similarity measures. 
In the \gls{t}, a computational measure of semantic similarity is seen as a human expert giving a judgement on the similarity of two given pairs.
Like human experts, similarity measures often disagree, and it is often difficult to identify unequivocally the best measure for a given context.
The ensemble approach is inspired by findings in risk management, machine learning, biology, and econometrics, which indicate that analyses that aggregate expert opinions from different experts tend to outperform analyses from single experts \cite{clemen:1999:combining,armstrong:2001:combiningforecasts,rokach:2010:ensembleclassif}.
Based on empirical results collected on WordNet-based similarity measures in the context of geographic terms, the following conclusions can be drawn:

\begin{itemize}

  \item An \gls{shortt} $E$, whose members are semantic similarity measures, is generally less cognitively plausible than the best of its members, i.e. $max( \rho_{sim} ) > \rho_E$. In $\approx 9\%$ of cases, the \gls{shortt} obtains total success, i.e. it outperforms the most plausible measure.
  The larger the \gls{shortt}, the less frequently the \gls{shortt} outperforms its best member.
    \item On average, similarity \glspl{shortt} $E$ tend to be more cognitively plausible than any of its individual measures $sim$ in isolation (mean of partial success ratio $\approx 70\%$).   In our evaluation, \glspl{shortt} with 3 members are the most successful. 
     \item The \gls{t} confirms what Cooke and Goossens \cite{cooke:2004:expert} pointed out in the context of risk assessment: ``a group of experts tends to perform better than the average solitary expert, but the best individual in the group often outperforms the group as a whole'' (p. 644).
    \item In the vast majority of cases ($\geq 98\%$), the cognitive plausibility of an \gls{t} is higher than the mean and median of its members' plausibilities.
    An \gls{shortt} is more plausible than the mean (or median) of its parts.
    These results are overall consistent with a preliminary evaluation \cite{Ballatore:2012:jury}.
  \item Individual similarity measures obtain widely different cognitive plausibility on different ground truths and contexts.
  In a context of limited information in which the optimal measure is unknown, we believe that the \gls{t} should be favour\-ed over any individual similarity measure.
\end{itemize}

\noindent Several issues should be considered for future work.
This study focused exclusively on ten WordNet-based similarity measures and, 
to gather more empirical evidence, the ensemble approach should be extended to different similarity measures.
Moreover, to aggregate the similarity scores, we have adopted two simple ensemble methods (the mean of scores and the mean of rankings).
More sophisticated ensemble techniques based on machine learning could be explored to increase the \gls{shortt}'s performance \cite{renda:2003:web}.
Furthermore, the empirical evidence presented in this paper was limited to the geographic context.
General-purpose semantic similarity datasets, such as that devised by Agirre \emph{et al.} \cite{agirre:2009:study}, could be used to further evaluate the \gls{shortt} across various semantic domains.

The evaluation utilised in this study is based on ranking comparison, which allows to quantify the cognitive plausibility of semantic similarity measure directly.
Although this approach is the most popular in the literature, it has several drawbacks, as extensively discussed by Ferrara and Tasso \cite{ferrara:2013:evalrelatedness}.
Alternatively, task-based evaluations could be used to assess the cognitive plausibility of measures indirectly by observing their ability to support a specific task.
Suitable tasks in \gls{gir} and \gls{nlp}, such as geographic query expansion, could be devised and deployed to evaluate the \gls{t} further.
In this study, similarity is modelled as a continuous score, but it can also be represented as a set of discrete classes.
More importantly, the evaluation discussed in this article focuses on \emph{acontextual} judgements of similarity of geographic terms.
Context, however, has been identified as a crucial component of similarity \cite{kessler:2007:similarity}, and the \gls{t} should extended to capture specific facets of the observed terms.
The effectiveness of the \gls{shortt} should be assessed when observing either the affordances, the size or the physical structure of geospatial entities.
 
The importance of semantic similarity measures in information retrieval, \gls{nlp}, and data mining can hardly be underestimated \cite{janowicz:2011:semantics,kuhn:2013:cognling}.
In this article, we have shown that a scientific contribution can be given not only by devising new similarity measures, but also by studying the combination of existing measures.
The \gls{t} provides a general approach to obtain more cognitively plausible results in settings where the ground truth is unstable and shifting.
 

\paragraph{Acknowledgements.}
The research presented in this paper was funded by a Strategic Research Cluster grant (07/SRC/I1168) by Science Foundation Ireland under the National Development Plan. The authors gratefully acknowledge this support.

\bibliography{thesis}

\end{document}

%% file: macros.tex
\newcommand{\tick}{$\surd$}

\newcommand{\footurl}[1]{\footnotemark\footnotetext{\url{#1}}} 
\newcommand{\footurltwo}[2]{\footnotemark\footnotetext{\url{#1}, \url{#2}}} 
\newcommand{\homepage}[1]{\url{http://github.com/ucd-spatial/#1}}

\newcommand{\nensembles}{1,012}

%% file: glossary.tex
\newacronym{t}{SSE}{semantic similarity ensemble}
\newglossaryentry{shortt}{name={ensemble}}
\newglossaryentry{kb}{name={knowledge base}}
\newglossaryentry{gkb}{name={geo-knowledge base}}
\newglossaryentry{gir}{name={geographic information retrieval}}
\newglossaryentry{osm}{name={OpenStreetMap}}
\newglossaryentry{wn}{name={WordNet}}

\newacronym{osmsim}{NetLexSiM}{Network-Lexical Similarity Measure}
\newacronym{vgi}{VGI}{Volunteered Geographic Information}
\newacronym{hci}{HCI}{human-computer interaction}
\newacronym{ai}{AI}{artificial intelligence}
\newacronym{ir}{IR}{information retrieval}
\newacronym{gis}{GIS}{Geographic Information Systems}
\newacronym{idf}{IDF}{Inverse Document Frequency}
\newacronym{lsa}{LSA}{Latent Semantic Analysis}
\newacronym{poipl}{POI}{points of interest}
\newacronym{poi}{POI}{point of interest}
\newacronym{vsm}{VSM}{Vector Space Model}

\newacronym{pos}{POS}{part-of-speech}
\newacronym{skos}{SKOS}{Simple Knowledge Organization System}
\newacronym{rdf}{RDF}{Resource Description Framework}
\newacronym{owl}{OWL}{Web Ontology Language}
\newacronym{giscience}{GIScience}{geographical information science}
\newacronym{wsddef}{WSD}{Word Sense Disambiguation}
\newacronym{dl}{DL}{description logic}
\newacronym{mdsmsim}{MDSM}{Matching-Distance Similarity Measure}
\newacronym{oursurvey}{GeReSiD}{Geo Relatedness and Similarity Dataset}
\newacronym{ira}{IRA}{interrater agreement}
\newacronym{bow}{BOW}{bag-of-words}
\newacronym{irr}{IRR}{interrater reliability}
\newacronym{mds}{MDS}{multidimensional scaling}
\newacronym{lbs}{LBS}{location-based services}
\newacronym{oss}{FOSS}{free and open-source software}
\newacronym{sdts}{SDTS}{Spatial Data Transfer Standard}
\newacronym{tfidf}{TF-IDF}{Term Frequency-Inverse Document Frequency}
\newacronym{ogc}{OGC}{Open Geospatial Consortium}

\newglossaryentry{w2}{name=Web 2.0,
	description={TODO}
}

\newglossaryentry{stratag}{name=Strategic Research in Advanced Geotechnologies (StratAG)}

\newglossaryentry{nuim}{name={National University of Ireland, Maynooth},
	description={TODO}
}

\newglossaryentry{ucd}{name={University College Dublin},
	description={TODO}
}

\newglossaryentry{lexsim}{name=\textsc{osm-sim_{lex}},
	description={TODO}
}

\newglossaryentry{simdl}{name=Sim-DL,
	description={TODO}
}

\newglossaryentry{netsim}{name=\textsc{osm-sim_{sim}},
	description={TODO}
}

\newglossaryentry{dbp}{name=DBpedia,
	description={TODO}
}

\newglossaryentry{lgd}{name=LinkedGeoData,
	description={TODO}
}

\newglossaryentry{sgw}{name=Semantic Geospatial Web,
	description={TODO}
}

\newglossaryentry{sw}{name=Semantic Web,
	description={TODO}
}

\newglossaryentry{webplatform}{name=Web platform for map personalisation and visualisation,
	description={TODO}
}

\newglossaryentry{wsd}{name=word sense disambiguation,
	description={TODO}
}

\newglossaryentry{os}{name=open source,
	description={TODO}
}

\newglossaryentry{osn}{name=OSM Semantic Network,
	description={TODO}
}

\newglossaryentry{nlp}{name=natural language processing,
	description={TODO}
}

\newglossaryentry{owc}{name=OSM Wiki Crawler,
	description={TODO}
}

\newglossaryentry{oww}{name=OSM Wiki website,
	description={\url{http://wiki.openstreetmap.org}}
}

\newglossaryentry{mdsm}{name=MDSM evaluation dataset,
	description={TODO}
}